*Mersin Üniversitesi Dil ve Edebiyat Dergisi, MEUDED, 2015; 12 (1): 43-61.*# ASSOCIATIVE MEASURES AND MULTI-WORD UNIT EXTRACTION IN TURKISH

Ümit Mersinli[1]

*Mersin University***Abstract**: Associative measures are "mathematical formulas determining the strength of association between two or more words based on their occurrences and cooccurrences in a text corpus" (Pecina, 2010, p. 138). The purpose of this paper is to test the 12 associative measures that Text-NSP (Banerjee & Pedersen, 2003) contains on a 10-million-word subcorpus of Turkish National Corpus (TNC) (Aksan et.al., 2012). A statistical comparison of those measures is out of the scope of the study, and the measures will be evaluated according to the linguistic relevance of the rankings they provide. The focus of the study is basically on optimizing the corpus data, before applying the measures and then, evaluating the rankings produced by these measures as a whole, not on the linguistic relevance of individual n-grams. The findings include intra-linguistically relevant associative measures for a comma delimited, sentence splitted, lower-cased, well-balanced, representative, 10-million-word corpus of Turkish.

**Keywords**: *Multi-word units, associative measures, Turkish National Corpus*[1] Mersin Üniversitesi, Fen Edebiyat Fakültesi, İngiliz Dili ve Edebiyatı Bölümü, umitmersinli@gmail.com
  Makale gönderim tarihi: 5 Mart 2015 ; Kabul tarihi: 2 Nisan 2015

# BİRLİKTELİK ÖLÇÜLERİ VE TÜRKÇEDE ÇOKSÖZCÜKLÜ BİRİM ÇIKARIMI

**Öz**: Birliktelik ölçüleri "bir dil derleminde, iki ya da daha fazla sözcük arasındaki ilinti gücünün, tek tek ve birlikte kullanımları temelinde belirlenmesinde kullanılan matematiksel formüllerdir" (Pecina, 2010). Bu makalenin amacı da Text-NSP (Banerjee & Pedersen, 2003) adlı yazılımın içerdiği 12 birliktelik ölçüsünü, Türkçe Ulusal Derlemi'nin (Aksan vd., 2012) veritabanlarından oluşturulan, 10 milyon sözcüklük bir alt-derlemde sınamaktır. Bu ölçülerin istatistik yöntemlerle karşılaştırılması çalışmanın kapsamı dışındadır. Bu yazı kapsamında sınanan birliktelik ölçüleri, oluşturdukları sıralamanın dil-içi uygunluğuna göre değerlendirilecektir. Çalışmanın odağında, istatistik ölçülerin uygulanması öncesinde derlem verisinin iyileştirilmesi ve ölçülerin uygulanması sonrasında oluşan sıralamaların, tek tek çok sözcüklü birimlerin uygunluğuna göre değil, sayısal sıralamanın dil-içi uygunluğa göre değerlendirilmesi vardır. Çalışmanın bulguları; virgülle sınırlanmış, tümcelerine ayrılmış, küçük harfe çevrilmiş, dengeli ve temsil yeterliği olan 10 milyon sözcüklük Türkçe bir derlem için dilbilimsel olarak uygun birliktelik ölçülerini içermektedir.

**Anahtar sözcükler:** *Çok sözcüklü birim, birliktelik ölçüsü, Türkçe Ulusal Derlemi*

## 1. INTRODUCTION

Jackendoff (1997) notes that the number of MWUs in a speaker's lexicon is of the same order of magnitude as the number of single words. Although we do not have any statistical estimation for the amount of MWUs in Turkish lexicon, the importance of these lexical units can be figured out with their proportion in English lexicon. For instance, as Ramisch et.al. (2013) notes, among the nouns in WordNet, 60.292 of the total 117,827 (51.4%) are MWUs and for the verbs, the proportion is 25.5% (2,829 among 11,558).

In this respect, there is a strong need for studies on MWU extraction in Turkish and the intra-linguistic properties of those MWUs, considering the fact that the overall proportion of MWUs in current Turkish lexicon appears to be greater than documented in dictionaries. Only after forming a preliminary, gold standard MWU set for Turkish, it could be possible for researchers to evaluate their statistical or linguistic methods and to improve the documentation of Turkish lexicon.

## 1.1. THEORETICAL BACKGROUND

Sinclair's (1991) "idiom principle" which asserts that we have a tendency to use MWUs, rather than storing and processing words individually, is the main assumption of this study. As Sinclair (ibid.) states, "the principle of idiom is that a language user has available to him or her a large number of semi-preconstructed phrases that constitute single choices, even though they might appear to be analysable into segments". Our definition of MWUs also follows Sinclair in the sense that they include not only well-known idioms as presented in current Turkish dictionaries but also other preconstructed "phrasemes" (Mel'čuk, 1995, p. 168).

In short, as Mel'čuk (1995, p. 169) states, "people speak not in words but in phrases" and for Turkish, even a preliminary, well-documented MWU lexicon has not been made available to public yet, which would also be a valuable resource for other rich-morphology languages.

## 1.2. SCOPE

The focus of this study is not on 'statistical' requirements, comparisons, optimizations or assessment of associative measures but on 'intra-linguistic, phraseological' relevance of the rankings provided by them, applied on a well-balanced, representative corpus of Turkish.Corpus-driven, directional (Evert, 2004) collocate extraction practices for pre-defined query words are off the scope of current study. It is limited to symmetrical n-gram rankings and their intra-linguistic validation for further studies in Turkish.

The study is limited to 2, 3 and 4-grams and the 12 associative measures in the package Text-NSP v1.27 (Banerjee & Pederson, 2003). Broader units or the identification of part-whole relationships between n and n+1 grams are excluded from the study.

Limitations of the software, if any (i.e. formulas, procedures or the code), are preserved as-is.

**2. CURRENT TRENDS IN MWU EXTRACTION**

Combining more than one association measure (Pecina, 2010) or applying each measure for a single grammatical pattern (Nissim & Zaninello, 2013) are the two current trends in MWU extraction. The later hybrid approach is highly adaptable to Turkish since it combines both grammatical filtering and statistical ranking. A classification of annotated n-grams according to their grammatical patterns, before applying any statistical measures, is proven to be productive for identifying MWUs, especially in a candidate set which consists of inflected forms of the same unit as the case for the agglutinative languages.

Rayson et.al. (2010, p. 2) also state that "it has become increasingly obvious that, in order to develop more efficient algorithms, we need deeper understanding of the structural and semantic properties of MWEs, such as morpho-syntactic patterns, semantic compositionality, semantic behaviour in different contexts, cross-lingual transformation of MWE properties etc.". As stated above, the development of efficient measures is only possible by focusing or operating on some morpho-syntactic classifications. Since associative measures are language-specific, experimenting on the measures applied on another language (mostly English) and expecting similar results is not a productive approach for Turkish.

**3. MWU EXTRACTION IN TURKISH**

Preliminary works on MWU extraction in Turkish - the first following a rule-based approach and the later a statistical one - are (Oflazer et.al., 2004) and (Kumova-Metin & Karaoğlan, 2010).

Oflazer et.al (2004) argue that MWUs –actually two-word MWUs- can be classified into *lexicalized*, *semi-lexicalized* and *non-lexicalized* units. However, since the extraction rules are based on a limited set, namely *light verb constructions* and *reduplications*, which are - by definition - the most frequent MWU forms in bigrams but not representative for further 3, 4-word MWUs, this classification is based on practical purposes rather than linguistic. In addition, the rules –although not documented properly- are designed to identify that limited set of grammatical patterns which are easier to define in regular expressions. We cannot solely rely on grammatical patterns in that sense, considering the varying internal, grammatical properties of MWUs, if not limited to bigrams.

Kumova-Metin and Karaoğlan (2010) state that Mutual Information and Chi-square are the two relevant measures for Turkish. This study, on the other hand, is strictly based on statistical measures to extract MWUs and thus ignores the occurrences of the same MWUs in different inflectional word forms by evaluating each inflectional variety as seperate units.

The present study will demonstrate the results of statistical measures on a comma delimited, sentence splitted, lower-cased corpus. The findings of the study, can also be optimized in further studies following a hybrid approach, which combines a grammatical classification as the first step and a relevant statistical ranking as the second.

Formulaicity in Turkish is also discussed in Durrant & Matthew-Aydınlı (2011) focusing on academic texts. As another genre-based approach, Aksan and Aksan (2013) provide a linguistic classification of MWUs in fiction and informative texts in terms of their grammatical patterns and discourse functions. This study is also valuable in being the first to have a representative, well-balanced corpus of Turkish as the data source. As a study on processes of symmetrical and directional MWU extraction, Mersinli and Demirhan (2012) demonstrate a case study on Primary School Turkish Language Teaching Coursebooks.

When the above mentioned studies are considered, it is apparent that hybrid approaches combining grammar-based filtering and statistical ranking will be the forthcoming trend in Turkish MWU extraction. Genre-based studies will also provide valuable findings and data for further studies.

## 4. DATA AND METHODOLOGY

4.1. OPERATIONAL DEFINITIONS

***Multi-Word Unit (MWU):*** The "non-compositional", "non-modifiable" and "non-substitutable" (Manning & Schütze, 2001, p. 184) word co-occurrences that are stored and processed as a single unit in mental lexicon. Since, the above mentioned characteristics do not have clear boundaries in all languages and MWUs are 'mostly' non-compositional, non-modifiable and non-substitutable, and also idiosyncratic; boundaries between a MWU and a syntactic phrase is highly dependent on the evaluater. As Calzolari (2002, p. 1934) states, MWUs "defy naïve attempts to establish a border between grammar and lexicon in terms of the opposition between rule productivity and lexical idiosyncrasy". Thus, in this study, the term 'multi-word unit' will refer to any fixed word sequences that can be processed as a single unit while preparing a dictionary. Wray (2002) discusses more than 50 terms referring to formulaic language use but discussing the varying terminological choices is out of the scope of current study.

***N-gram:*** all co-occurrences –either lexicalized or not- of two or more words recurring in a corpus and extracted from a corpus as MWU candidates. N refers to the number of words included.

***Word:*** A "fuzzy", "language-specific" (Haspelmath, 2011) concept which cannot be an operational unit cross-linguistically. A given word in a language can be an affix in another language, as the case for English and Turkish. Thus, in this study, "word" refers to any sequence of characters defined as tokens in a corpus, in other words, an ortographical unit rather than a linguistic one, delimited with space characters or certain punctuation marks.

***Associative measures:*** "Mathematical formulas determining the strength of association between two or more words based on their occurrences and cooccurrences in a text corpus" (Pecina, 2010, p. 138). These formulas provide n-gram rankings as an initial step for MWU extraction.

4.2. THE CORPUS

The data of the study is derived from a 10-million-word sub-corpus (TNC-Baby) following the design features of Turkish National Corpus (http//www.tnc.org.tr) and covering a period of 20 years (1990-2009). The sub-corpus preserves the quantificational distribution of TNC in terms of text domains, time and medium of texts.

Textual data in the sub-corpus is optimized for practical purposes and the ASCII formatted, sentence splitted, comma delimited, lower-cased text is processed on a Windows PC with Turkish as the system language, and Perl v.5.16.2. Table 1 presents the optimization process. Excluding the punctuation marks as non-tokens is a well-known practice in similar studies. Our proposal is that noise-reduction or excluding ill-formed MWU candidates should also be included in the pre-formatting of corpus data, before applying any measure. A comma delimited text, for instance, will exclude irrelevant ngrams and may provide less noisy rankings.

**Table 1.** Optimizing corpus data for MWU extraction

**original, sentence-splitted text**

> Ekoloji
> YAPRAK DÖKÜNTÜLERİNDE FUNGAL SUKSESYON
> Bu makalede, çam yaprakları ve diğer ağaç yapraklarının çürümeleri anlatılmıştır.

**optimized version (lower-cased, comma delimited)**

> ekoloji
> yaprak döküntülerinde fungal suksesyon
> bu makalede
> çam yaprakları ve diğer ağaç yapraklarının çürümeleri anlatılmıştır.

4.3. SOFTWARE

The Perl package Text-NSP v1.27 is used to rank the n-grams according to their observed frequency (count.pl) and to compute 12 associative measures (statistic.pl). For practical purposes, the line 'use locale,' is added to the Perl code, both for count.pl and statistic.pl modules, to handle Turkish-specific characters in ASCII formatted corpus data. Table 2 summarizes the usage of the package, where 'ignore.txt' includes the punctuation marks not to be regarded as tokens and 'remove 10' declares the cut-off point simply to exclude n-grams occurring less than 10 times in the corpus.

**Table 2.** Usage of Text-NSP v1.27

```
perl count.pl -ngram 2 -nontoken ignore.txt -newLine -remove 10 output1.count corpus.txt
perl statistic.pl --ngram 2 Text::NSP::Measures::2D::CHI::tscore output2.txt output1.count
```

4.4. ASSOCIATIVE MEASURES

The 12 measures in Table 3 are the set of associative measures evaluated.

**Table 3.** The Measures of Association Provided in Text-NSP v.1.27

| 2-gram measures | Abbreviation |
|---|---|
| Dice Coefficient | dice |
| Fishers exact test - left sided | left |
| Fishers exact test - right sided | right |
| Fishers exact test - two tailed | twotailed |
| Jaccard Coefficient | jaccard |
| Log-likelihood ratio | ll |
| Mutual Information | mi |
| Pointwise Mutual Information | pmi |
| Phi Coefficient | phi |
| Pearson's Chi Squared Test | x2 |
| Poisson Stirling Measure | ps |
| T-score | tscore |

| 3-gram measures | Abbreviation |
|---|---|
| Log-likelihood ratio | ll |
| Mutual Information | mi |
| Pointwise Mutual Information | pmi |
| Poisson Stirling Measure | ps |

| 4-gram measures | Abbreviation |
|---|---|
| Log-likelihood ratio | ll |

4.5. EVALUATION OF THE MEASURES

The relevance of associative measures for n-gram ranking is often measured statistically by calculating precision, recall or f-measure (the combination of precision and recall) values which is not the preferred technique in this study. The rationale for this preference is that, besides being data or language specific, those values rely on whether the ranked n-grams are valid MWUs or not, which is problematic for Turkish having no standard, representative MWU datasets published for such validation.

In addition, according to Pecina (2010), eliciting the best association measure for MWU extraction depends heavily on data, language, and the notion of MWU itself. As Hiemstra & Kraaij (2007, p. 356) state, "it takes more discipline to perform a really blind experiment and extra care not to tune on the -*statistical*- data".

MWU extraction is a semi-automatic, corpus-based Natural Language Processing (NLP) practice that includes ranking and filtering n-grams which validate pre-defined statistical and/or linguistic criteria. Those pre-defined criteria leads to data modification to some extent (e.g.

thresholds or the reference MWU sets) and thus prevent the MWU extraction studies from being fully automatic processes without any intervention by the researcher.

To summarize, our argument for the evaluation of associative measures is that any measure can be considered as valid, depending on the following criteria.

i. the purpose
> (i.e., extracting Light Verb Constructions, Named Entities or Discourse Connectives, Postpositional Phrases, Genitive-Possessive constructions, Compound Nouns etc or even extracting non-MWUs)

ii. the language (e.g., isolating or agglutinative)
> (i.e., a word in an isolating language would possibly be the equivalent of a suffix and thus the measures themselves are again language-specific)

iii. unit of interest
> (i.e. types, lemmas, suffixes or their combinations)

iv. data
> (i.e. a well-balanced, representative reference corpus versus a genre-specific web-based text archive)

v. statistical data modifications
> (i.e. setting the optimum thresholds for high precision, recall values)

With respect to the above mentioned considerations, our evaluation will be based on a single validation unit for 2-grams, namely *ya da* "or", a variant of *veya* in Turkish. That single MWU serves as an evaluater for the rankings of the measures subject to the study provided, because, without doubt, any ranking should start with that MWU if the data source is representative for Turkish, especially considering the finding that top-rated n-gram provided by the measures evaluated is either *ya da* "or" or *teker teker* "one by one". This distinction makes the rankings starting with *teker teker,* irrelevant for general MWU extraction, regarding the huge difference in the observed frequency values of the two.

The rationale behind using such a positive evidence for valid MWU rankings is just similar to the use of stop words that cannot initialize or finalize a MWU in Turkish – e.g. *ve* "and", *de, da* "too" or *bir* "a, an" - as a negative evidence for identifying false positives within a ranking.

On the other hand, for 3-grams, *ya da* "or", serves as a negative evidence for invalid rankings, since any relevant ranking should not start with 3-grams with an initial *ya da* "or" sequence as in Table 4.

**Table 4.** Examples for irrelevant 3-gram rankings

|    | **Log-likelihood** | **True mutual information** |
|----|---------------------|------------------------------|
| 1  | ya da bu            | ya da bu                     |
| 2  | ya da başka         | ya da başka                  |
| 3  | ya da böyle         | ya da böyle                  |
| 4  | ya da olumsuz       | ya da olumsuz                |
| 5  | ya da benim         | ya da kişisel                |
| 6  | ya da kişisel       | ya da benim                  |
| 7  | ya da ne            | ya da daha                   |
| 8  | ya da daha          | ya da ne                     |
| 9  | ya da diğer         | ya da diğer                  |
| 10 | ya da onun          | ya da onun                   |
| 11 | ya da bana          | ya da en                     |
| 12 | ya da birkaç        | ya da birkaç                 |
| 13 | ya da en            | ya da bana                   |
| 14 | ya da kendi         | ya da yeni                   |
| 15 | ya da yeni          | ya da kendi                  |
| 16 | ya da dolaylı       | ya da çok                    |
| 17 | ya da her           | ya da hiç                    |
| 18 | ya da çok           | ya da her                    |
| 19 | ya da yanlış | ya da özel |
| 20 | ya da özel | ya da dolaylı |

For 4-grams, the evaluation is based on no pre-defined stop-lists since log-likelihood is the only associative measure applicable to 4-grams in Text-NSP. Thus, only an overall evaluation of top-20 4-grams ranked by that measure will be presented in the study.

**5. FINDINGS AND DISCUSSION**

5.1. BIGRAMS

Only 5 of the 12 associative measures, namely T-score, Fisher's exact test (left-sided), Log-likelihood, True Mutual Information and Poisson-Stirling, provided valid rankings starting with *ya da* "or" for bigrams. Table 5 presents the top-20 ranked MWU candidates for each measure. Abbreviations for the measures are the ones stated in Table 3.

**Table 5.** Valid associative measures for 2-grams and top-20 MWU candidates.

| | tscore | left | ll & tmi | ps |
|---|---|---|---|---|
| 1 | ya da | ya da | ya da | ya da |
| 2 | hem de | ve bu | söz konusu | söz konusu |
| 3 | bir şey | bir şey | hem de | hem de |
| 4 | ne kadar | hem de | bir bir | aynı zamanda |
| 5 | böyle bir | böyle bir | aynı zamanda | olmak üzere |
| 6 | söz konusu | büyük bir | olmak üzere | ne kadar |
| 7 | büyük bir | ne kadar | bir ve | ile ilgili |
| 8 | bu nedenle | önemli bir | ne kadar | yer alan |
| 9 | başka bir | başka bir | ile ilgili | yanı sıra |
| 10 | ben de | yeni bir | yer alan | öte yandan |
| 11 | daha çok | bir şekilde | yanı sıra | daha fazla |

| 12 | önemli bir | daha çok | öte yandan | son derece |
|----|------------|----------|------------|------------|
| 13 | aynı zamanda | söz konusu | bu nedenle | değil mi |
| 14 | ile ilgili | ben de | böyle bir | bu nedenle |
| 15 | daha fazla | bu nedenle | daha fazla | olduğu gibi |
| 16 | o zaman | bir başka | son derece | pek çok |
| 17 | yeni bir | herhangi bir | ve ve | ortaya çıkan |
| 18 | olduğu gibi | daha sonra | değil mi | en önemli |
| 19 | olmak üzere | bir süre | belki de | sık sık |
| 20 | herhangi bir | o zaman | bir şey | belki de |

Almost all previous studies on MWU extraction in Turkish focus basically on 2-grams. The number of associative measures that are applicable to bigrams make them interesting for statistical reasons. However, 3-grams are more productive for MWU formation in Turkish since their morpho-syntactic content includes all the required components - i.e. heads, modifiers and specifiers - to form a closed syntactic projection. Linguistically speaking, as stated in Aksan & Aksan (2013), 3-grams are more relevant candidates for MWU extraction in Turkish. Thus, statistically oriented discussions or evaluations based on 2-grams and excluding further units make those findings specific to 2-grams, not for Turkish MWUs in general.

Another problem, more specific to 2-grams is that most of them are parts of larger units in Turkish, mostly 3-grams. Those fragmental 2-grams form the majority of the noisy data. However, it's a cross-linguistic problem, especially for 2-gram MWU candidates and substring reduction is another current trend in MWU extraction studies. As O'Donnell (2011, p. 136) states

> "A common methodological step in a corpus linguistic analysis is the extraction of frequency lists of various size chunks (variously called clusters, lexical bundles or n-grams). Most software packages facilitate the creation of such lists, making it possible to compare units of different length. However, each size unit is (necessarily) counted on its own terms without reference to larger units of which they may be a part."

In addition, ignoring part-whole relationships in any frequency-based lexical data is not a problem specific to MWU extraction. All word frequencies, for instance, also include the occurrences of those words in broader MWUs. In this respect, reducing the noise caused by fragmental data at any level in lexical statistics is a broader problem and thus, out of the scope of current study.

As stated in Section 4.5., our primary argument, following Pecina (2010), is that any measure can be relevant depending on your purpose. In this sense, all the evaluations above are about 2-grams for 'general' MWU extraction in Turkish.

We argue that invalid associative measures for 2-grams can also be used for specific purposes. For example, the measures Dice coefficient, Jaccard, Phi Coefficient and Pearson's Chi Squared Test provided identical rankings with reduplications on top. More specifically, 92 of the top 200 bigrams ranked by those measures are reduplications. Table 6 presents the top-50 reduplications and their rankings with the above measures.

**Table 6.** Reduplications ranked by Dice coefficient, Jaccard, Phi Coefficient and Pearson's Chi Squared Test

| |
|---|
| teker teker (1), irili ufaklı (4), peş peşe (5), kayıtsız şartsız (6), uçsuz bucaksız (7), apar topar (9), ışıl ışıl (10), cıvıl cıvıl (14), koşa koşa (16), seve seve (17), burun buruna (18), tir tir (19), doya doya (24), gürül gürül (25), cık cık (27), gizliden gizliye (28), boşu boşuna (29), omuz omuza (30), hüngür hüngür (31), topu topu (34), vah vah (35), içli dışlı (36), sağda solda (37), allak bullak (38), harıl harıl (40), kuşaktan kuşağa (41), kesik kesik (42), körü körüne (43), diri diri (48), mışıl mışıl (49), enine boyuna (54), haşır neşir (55), didik didik (56), kıpır kıpır (57), inceden inceye (61), canla başla (65), kıs kıs (68), tıkır tıkır (76), aşağıdan yukarıya (78), bitmez tükenmez (80), vura vura (81), abuk sabuk (82), iner inmez (83), dalgın dalgın (84), derme çatma (87), kıran kırana (88), cayır cayır (90), döne döne (93), oluk oluk (98), havadan sudan (99) |

Finally, Fishers exact test (right sided and two tailed) provided irrelevant rankings in contrast with left-sided Fisher's and are worth considering for future comparisons or identifying non-MWUs.

5.2. TRIGRAMS

The only valid 3-gram rankings for general MWU extraction in Turkish is provided by Poisson-Stirling measure. As presented in Section 3.5., Log-likelihood and True Mutual Information rankings, starting all with *ya da* "or" are considered to be invalid for general MWU extraction from 3-gram candidates. Table 7 compares the Poisson-Stirling rankings with the observed frequencies of 3-grams. Regarding the fact that most of the MWUs in Turkish can be extracted from 3-grams, and the limitations of statistical measures, we can argue that observed frequencies are also valuable at 3-gram level.

**Table 7.** Valid rankings for 3-grams and top-20 MWU candidates.

|    | **observed freq.** | **ps** |
|----|--------------------|--------|
| 1  | bir süre sonra     | ne var ki |
| 2  | bir kez daha       | ne yazık ki |
| 3  | ne var ki          | her ne kadar |
| 4  | her ne kadar       | bir kez daha |
| 5  | başka bir şey      | ne olursa olsun |
| 6  | ne yazık ki        | bir süre sonra |
| 7  | bir yandan da      | her şeyden önce |
| 8  | çok önemli bir     | başka bir şey |
| 9  | bir an önce        | bir an önce |
| 10 | ne olursa olsun    | başta olmak üzere |
| 11 | kısa bir süre      | kısa bir süre |
| 12 | her şeyden önce    | bir yandan da |
| 13 | ya da bir          | radyo ve televizyon |
| 14 | başka bir deyişle  | ses kalitesi okuma |
| 15 | çok büyük bir | ile ilgili olarak |
| 16 | daha önce de | buna bağlı olarak |
| 17 | bir başka deyişle | dahil olmak üzere |
| 18 | böyle bir şey | her geçen gün |
| 19 | ile ilgili olarak | ama yine de |
| 20 | ya da bu | daha önce de |

5.3. FOUR-GRAMS

Below are the Log-likelihood rankings and the observed frequencies of 4-gram MWU candidates. Due to the number of measures applicable for 4-grams, stop-word filtering (e.g. "ve", "da") can be used as the first step for general MWU extraction from 4-gram candidates.

**Table 8.** Top-20 MWU candidates ranked by log-likelihood and observed frequencies.

| | **Raw** | **ll** |
|---|---|---|
| 1 | kısa bir süre sonra | ve bir süre sonra |
| 2 | başka bir şey değildir | kısa bir süre sonra |
| 3 | şu ya da bu | kısa bir süre için |
| 4 | ve buna bağlı olarak | kısa bir süre önce |
| 5 | her zaman olduğu gibi | kısa bir süre içinde |
| 6 | petrol ve doğal gaz | başka bir şey değildir |
| 7 | bir o kadar da | bir ya da iki |
| 8 | g e g için | bir ya da birkaç |
| 9 | başbakan recep tayyip erdoğan | ama bir süre sonra |
| 10 | de dahil olmak üzere | kısa bir süre içerisinde |
| 11 | türkiye büyük millet meclisi | bir ya da birden |
| 12 | iş doyumu ve yaşam | da bir süre sonra |
| 13 | ama ne yazık ki | de bir süre sonra |
| 14 | ne var ki bu | belli bir süre sonra |
| 15 | küçük ve orta ölçekli | şu ya da bu |
| 16 | özel radyo ve televizyon | başka bir şey yok |
| 17 | doyumu ve yaşam doyumu | ve bir o kadar |
| 18 | kısa bir süre önce | ve bir kez daha |
| 19 | ya da başka bir | geçici bir süre için |
| 20 | çok kısa bir süre | başka bir şey değildi |

## 6. CONCLUSION

Table 9 summarizes the associative measures validated linguistically for general MWU extraction in Turkish.

**Table 9.** Valid associative measures for general MWU extraction in Turkish

| | |
|---|---|
| 2-grams | T-score, Fisher's Exact Test (left-sided), Log-likelihood, True Mutual Information, Poisson-Stirling Measure |
| 3-grams | Poisson-Stirling Measure |
| 4-grams | Log-likelihood |

According to the findings above and considering the fact that 3-grams should be of special interest for MWU extraction in Turkish, we can argue that associative measures should be used with a preceding or following grammatical filtering stage and a hybrid approach combining rule based and statistical techniques is a necessity if not a must in Turkish. As discussed in Aksan et.al. (2015), a grammatical classification before applying any statistical measure, can exclude most of the non-MWUs from an n-gram ranking, since they will not validate the morphosyntactic constraints as presented in (1-2).

(1) ADJECTIVE_DETERMINER_NOUN

*kısa bir süre* "a short time"

*belli bir süre* "a limited time)

(2) * PRONOUN_NOUN+loc_NOUN

*bu sınavda başarı* "success in this exam"

*bu konuda fikir* "argument on this topic"

REFERENCES

Aksan, M. & Aksan, Y. (2013). *Multi-word units and pragmatic functions in genre specification.* Paper presented at 13th IPrA Conference 08-13 September 2013. New Delhi, India.

Aksan, Y. et al. (2012). Construction of the Turkish National Corpus (TNC). *Proceedings of the Eight International Conference on Language Resources and Evaluation (LREC 2012).* (pp. 3223-3227) İstanbul. Turkiye.

Aksan, Y., Mersinli, Ü. & Altunay, S. (2015). *Colligational analysis of Turkish multi-word units.* Paper presented at CCS-2015, Corpus-Based Word Frequency: Methods and Applications. 19-20 February 2015. Mersin University. Turkiye.

Banerjee, S & Pederson, T. (2003). The design, implementation and use of the (N)gram (S)tatistic (P)ackage. *Proceedings of the Fourth International Conference on Intelligent Text Processing and Computational Linguistics*, (pp. 370-381).

Blanken, H.M., de Vries, A.P., Blok, H.E & Feng, L. (Eds). (2007). *Multimedia retrieval*. City: Springer.

Calzolari, N. et.al. (2002). Towards best practice for multiword expressions in computational lexicons. *Proceedings of the 3rd International Conference on Language Resources and Evaluation (LREC 2002).* (pp. 1934-1940 ). Las Palmas, Canary Islands.

Durrant, P. & Mathews-Aydınlı, J. (2011). A function first approach to identifying formulaic language in academic writing. *English for Specific Purposes, 30*, 58-72.

Evert, S. (2004). *The statistics of word cooccurrences (word pairs and collocations).* (Doctoral Dissertation). Universitat Stuttgart.

Jackendoff, R. (1997). *The architecture of the language faculty*. Cambridge, MA. MIT Press.

Haspelmath, M. (2011). The indeterminacy of word segmentation and the nature of morphology and syntax. *Folia Linguistica, 45*(1), 31-80.

Hiemstra, D. & Kraaij, W. (2007). Evaluation of multimedia retrieval systems. In H.M. Blanken, A.P. de Vries, H. E. Blok, L. Feng (Eds.), *Multimedia Retrieval* (pp. 347-366). Berlin: Springer.

Kumova-Metin, S. & Karaoğlan B., (2010). *Collocation extraction in Turkish texts using statistical methods.* Paper presented at 7th International Conference on Natural Language Processing (LNCS-ISI) IceTAL 2010, Reykjavik, Iceland.

Manning, C. D., & Schütze, H. (2001). *Foundations of statistical natural language processing*. Cambridge, Mass.: MIT Press.

Mel'čuk, I. A. (1995). Phrasemes in language and phraseology in linguistics. In .M. Everaert, E.-J. van der Linden, A. Schenk & R. Schreuder (Eds*.), Idioms: Structural and psychological perspectives* (pp. 167–232). Hillsdale, NJ: Lawrence Erlbaum.

Mersinli, Ü. & Demirhan, U. (2012). Çok sözcüklü kullanımlar ve ilköğretim Türkçe ders kitapları. In M. Aksan & Y. Aksan. (Eds.), *Türkçe Öğretiminde Güncel Çalışmalar.* (pp. 113-122). Mersin: Mersin Üniversitesi.

Nissim, M. & Zaninello, A. (2013). Modeling the internal variability of multiword expressions through a pattern-based method. *ACM Transactions on Speech and Language Processing*, *10*(2), 1-26.

O'Donnell. M.B. (2011). The adjusted frequency list: A method to produce cluster sensitive frequency lists. *ICAME Journal. Computers in English Linguistics*, 35, 135-169.

Oflazer, K., Çetinoğlu, Ö. & Say, B. (2004). Integrating morphology with multi-word expression processing in Turkish. *Proceedings of the Workshop on Multiword Expressions: Integrating Processing (MWE '04). Association for Computational Linguistics,* (pp.64-71). Stroudsburg, PA, USA.

Pecina, P. (2010). Lexical association measures and collocation extraction. *Language Resources and Evaluation, 44*(1-2), 137-158.

Ramisch, C., Villavicencio, A., & Kordoni, V. (2013). Introduction to the special issue on multiword expressions: From theory to practice and use. *ACM Transactions on Speech and Language Processing, 10*(2), pp. 1-3.

Sinclair, J.M., & Daley, J.S. (2004). *English collocation studies: The OSTI Report*. London, New York: Continuum.

Rayson, P., Piao, S., Sharoff, S., Evert, S., & Moiron, B.V. (2010). Multiword expressions: Hard going or plain sailing? *Language Resources and Evaluation, 44*(1-2), 1-5.

Wray, A. (2002). *Formulaic language and the lexicon*. Cambridge: Cambridge University Press.